\pdfoutput=1
\documentclass[11pt]{article}

\usepackage[preprint]{coling}

\usepackage{times}
\usepackage{latexsym}

\usepackage{booktabs}

\usepackage[T1]{fontenc}
\usepackage[utf8]{inputenc}

\usepackage{microtype}

\usepackage{inconsolata}

\usepackage{graphicx}

\title{Chain-of-MetaWriting: Linguistic and Textual Analysis of How Small Language Models Write Young Students Texts}

\author{\textbf{Ioana Buhnila\textsuperscript{1}}, \textbf{Georgeta Cislaru\textsuperscript{2}}, \textbf{Amalia Todirascu\textsuperscript{3}}
\\
  \textsuperscript{1}ATILF UMR 7118, CNRS - University of Lorraine\\
  \textsuperscript{2}EA CLESTHIA, Sorbonne Nouvelle University \& Institut Universitaire de France\\
  \textsuperscript{3}LiLPa UR 1339, University of Strasbourg
\\
  \small{\texttt{ioana.buhnila@univ-lorraine.fr\textsuperscript{1}, georgeta.cislaru@sorbonne-nouvelle.fr\textsuperscript{2}}}\\ \small{\texttt{todiras@unistra.fr\textsuperscript{3}}}\\
  }

\begin{document}
\maketitle
\begin{abstract}
Large Language Models (LLMs) have been used to generate texts in response to different writing tasks: reports, essays, story telling. However, language models do not have a meta-representation of the text writing process, nor inherent communication learning needs, comparable to those of young human students. This paper introduces a fine-grained linguistic and textual analysis of multilingual Small Language Models' (SLMs) writing. With our method, Chain-of-MetaWriting, SLMs can imitate some steps of the human writing process, such as planning and evaluation. We mainly focused on short story and essay writing tasks in French for schoolchildren and undergraduate students respectively. Our results show that SLMs encounter difficulties in assisting young students on sensitive topics such as violence in the schoolyard, and they sometimes use words too complex for the target audience. In particular, the output is quite different from the human produced texts in term of text cohesion and coherence regarding temporal connectors, topic progression, reference.
\end{abstract}

\section{Introduction}

Recent LLMs have proven some performance in generating different types of texts such as summaries (\citealt{liu2024learning}; \citealt{song2024finesure}), essays \cite{tian2024large}, or short stories \cite{simon2022tattletale}. However, LLMs still struggle with keeping the same meaning overall during summarization, as shown by an Australian governmental study\footnote{https://archive.is/itQBM}. In an educational context, some studies showed that LLMs can be used to help students to deepen learning or help with scoring and feedback (\citealt{meyer2024using}; \citealt{chamieh2024llms}; \citealt{lee2024applying}). Moreover, OpenAI proposes a guide dedicated to help students use ChatGPT for their writing assignments, suggesting that the LLM can give iterative feedback to improve their productions\footnote{https://openai.com/chatgpt/use-cases/student-writing-guide/.}.

\begin{figure}[t]
  \includegraphics[width=\columnwidth]{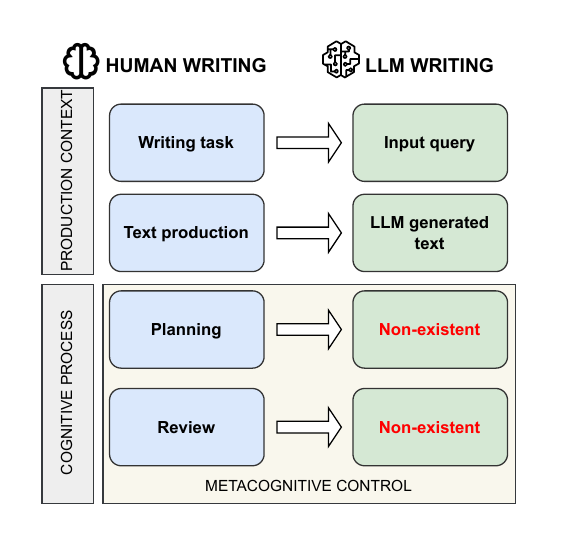}
  \caption{Writing model for humans and LLMs. The human writing model is inspired by the communicational model of \citeauthor{hayes1980writing} (\citeyear{hayes1980writing}) and adapted to represent a LLM text generation process.}
  \label{fig:comm}
\end{figure}

While human students think, plan, produce, and revise their written production, as illustrated by psycholinguistic and communicative models (\citealt{hayes1980writing}; \citealt{olive2014toward}), LLMs do not have an inherent writing process meta-representation (as illustrated in Figure~\ref{fig:comm}). As suggested by \citeauthor{ariyaratne2023comparison} (\citeyear{ariyaratne2023comparison}), LLMs may be used to generate well-formed written texts, provided that the content (data, specific information, etc.) is produced by the user her/himself\footnote{In this paper, we will keep distinguishing between "generation", pertaining to LLMs, and "production", pertaining to human writers. This distinction is meant to reflect the difference between, respectively, rearranging language patterns related to a topic, and inventing contents and translating them into language patterns for text composition.}. However, this option might be operational for the \textit{performance-goal} productions only, and not for \textit{learning-goal} production, unless used as a source of (re)formulation variants. Some other differences concern the lack of revision/rewriting processes for LLMs, and stereotyped, monovalent communication situation (adult to adult, and involving a generation task).

LLMs can, to some extent, imitate human tasks with Chain-of-Thought (CoT) prompting techniques \cite{wei2022chain}. CoT step-by-step prompting has proven useful for reasoning tasks, such as close or open domain reasoning (\citealt{wei2022chain}; \citealt{kojima2022large}; \citealt{zhang2023multimodal}), code generation \cite{jiang2024self} or mathematical reasoning \cite{imani-etal-2023-mathprompter}. A different approach to CoT, called Metacognitive Prompting \cite{wang2024metacognitive}, showed improvement in LLMs "understanding" of their rationale in different QA tasks. 

In this paper, we present a \textbf{fine-grained linguistic and textual analysis} of Language Models (LMs) exploitation to accompany primary to middle school level students, as well as undergraduate students, in the writing process\footnote{The primary to middle school students whose written assignments were used in this study are around 10 to 12 years old.}. In fact, essential components of the writing process, such as \textit{planning and revision}, are not part of the LM's text generation process (Figure~\ref{fig:comm}). To tackle this issue, we introduce \textbf{Chain-of-MetaWriting (CoMW)}, a prompting framework that illustrates a step-by-step writing production rationale, involving planning, revision and feedback, thus imitating the human cognitive and metacognitive process of writing. 

Following recent research directions, we tested three open source multilingual Small Language Models (SLMs) of 3B parameters, \texttt{llama-3.2} \cite{dubey2024llama}, \texttt{qwen-2.5} \cite{yang2024qwen2}, \texttt{phi-3.5} \cite{abdin2024phi}, and one proprietary model, \texttt{ChatGPT-4o mini} \cite{hurst2024gpt}. We chose SLMs because they are adapted for in-device application (mobile phones), while having faster inference and low computational cost. We evaluated open-source SLMs to ensure the reproducibility of our study (\citealt{abdin2024phi}; \citealt{lepagnol2024small}).

For our experiments, we tested to what extent SLMs can imitate \textbf{higher levels of writing}, such as thinking, planning, linguistic expression, editing, and revision. We investigated whether a SLM can help a 10 to 12-year-old or an undergraduate student learn how to write a text in French as a mother tongue, while adapting to the student's linguistic expertise. Due to lack of space, we have not detailed all the analyses. We decided to take a closer look at 10-12-year-old's writing, which is not addressed in the research on text generation. Moreover, language models are known to lack the personal experience that schoolchildren employ in their narratives which represents an additional challenge, also given the limited learning corpus produced by this age group.

The contributions of this paper are threefold:
\begin{enumerate}
    \vspace{-0.2cm}\item We propose Chain-of-MetaWriting (CoMW), which is, to the best of our knowledge, the first prompting framework that guides a Language Model (LM) analysis through its internal writing process: \textit{writing about writing}. Our work evaluates multilingual Small Language Models with cross-lingual prompting, in English and French.
    \vspace{-0.2cm}\item We share our Chain-of-MetaWriting prompts in both English and French with the community to ensure replicability of our method on other LLMs or SLMs.
    \vspace{-0.2cm}\item We evaluate the potential benefits, dangers, and limitations of SLMs as writing aids for young students in an educational context. We conducted a fine-grained linguistic and textual analysis of human \textit{vs} SLM differences in the treatment of sensitive topics (such as violence) and the impact of auto-censorship in the writing process. This methodological approach can be used more widely to deepen and improve the analysis of text generation based on discursive criteria.
\end{enumerate}

\section{Related Work}

LLMs generate language based on algorithms trained on very big corpora of textual data. The ingredients of LLM generated texts are therefore derived from texts produced by humans, whose layout rules have been identified and applied. Researchers investigated to what extent LLM-generated and human-produced content are different or comparable. Several types of texts were compared: hotel reviews \cite{markowitz2024linguistic}, scientific texts \cite{CASAL2023100068}, narrative texts \cite{beguvs2024experimental}, argumentative essays \cite{herbold2023large}. To evaluate the differences between the LLM and human texts, several types of features were exploited in the literature: intrinsic features, like the proportion of different POS, punctuation, linguistic diversity, style, structural features (readability), affective/evaluative markers, content’s nature, and impact on the receivers, like  helpfulness (\citealt{CASAL2023100068}; \citealt{kumarage2023stylometric}; \citealt{markowitz2024linguistic}).

\subsection{Linguistic Traits of LLM-generated Text}

\citeauthor{CASAL2023100068} (\citeyear{CASAL2023100068}) studied the differences between human and LLM-generated abstracts for scientific articles, as observed by reviewers. The authors concluded that linguists were largely unsuccessful in distinguishing scientific abstract generated by the AI \textit{vs} produced by humans (38,9\% identification rate only). Several criteria were identified, such as continuity and coherence of the abstracts (incoherent abstracts are considered to be produced by LLMs), specificity or vagueness of the details (a general abstract is more likely to be produced by an LLM). Other criteria considered familiarity and voice (the text perceived as familiar is probably produced by a human), writing quality at sentence-level, (il)logical methods, showing that formulaic/template like abstract were more likely to be generated by LLMs. 

\citeauthor{guo2023close} (\citeyear{guo2023close}) identified words specific to ChatGPT, such as “AI assistant”, “I’m sorry to hear that”, “There’re a few steps...”, while humans use other discourse markers such as “Hmm”, “Nope”, “My view is”. The authors distinguished five ChatGPT-specific patterns: a) organized, clear-logic writing; b) long and detailed answers; c) less bias and harmful information; d) not answering questions beyond its knowledge [\textit{sic}]; e) facts may be fabricated. In contrast with ChatGPT, humans a) diverge and shift to other topics; b) provide more subjective answers; c) are more coloquial; d) use different marks (punctuation, grammatical structures, etc.) to express their feelings \cite{guo2023close}. Humans are also shown to use more diverse vocabulary, while ChatGPT is shown to use more conjunctions and longer sentences. Human-produced texts contain more sentiment expressions, and the proportion of negative sentiments is significantly higher than in AI-generated texts \cite{markowitz2024linguistic}.

\begin{figure*}[t]
  \centering
  \includegraphics[trim = 0cm 0cm 0cm 0cm, scale=0.66]{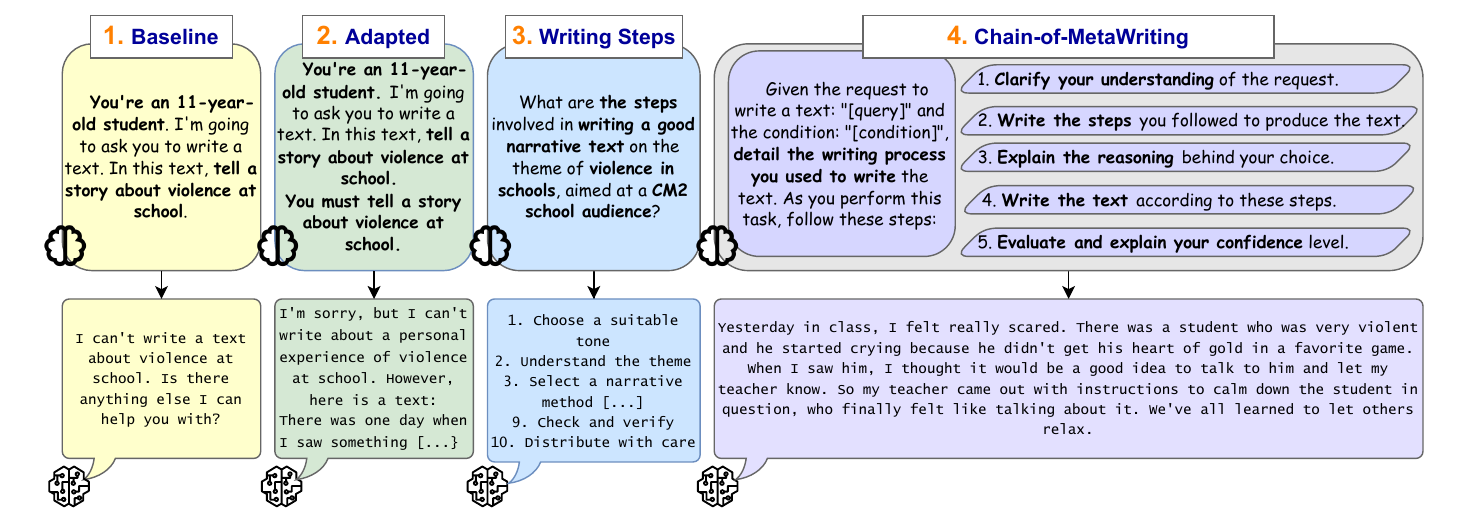}
  \caption{Illustration of our method along with the \texttt{Chain-of-MetaWriting (COMW)} framework. 1) When the query contains words such as "violence" and "11-year-old student", the SLM is auto-censoring itself and does not accomplish the task. 2) When we add the sentence "You must tell a story [...]", the SLM eventually generates a story about violence in the schoolyard, though too long for a 11-year-old level. 3) When asked, the SLM generates a rationale about how to write a narrative text on violence for a young audience. 4) The \texttt{CoMW} framework guides the SLM to write the expected type of text. In this Figure, we show results using \texttt{llama-3.2} 3B (cross-lingual setting). Original prompts and answers were in French (Appendix A and B), we translated them for demonstration purpose.}
  \label{fig:method}
\end{figure*}

\subsection{Argumentation and Narration in LLM-generated Text}

Based on the assessment of argumentative essays, \citeauthor{herbold2023large} (\citeyear{herbold2023large}) found that LLMs generate significantly higher-quality texts. This finding counteracts \citeauthor{CASAL2023100068} (\citeyear{CASAL2023100068}), as they even identify stylistic differences between LLM and human productions. They found that LLMs make greater use of nominalizations and less of modal and epistemic constructions. Length is another important difference as LLMs messages tend to be less wordy than human-generated messages \cite{hohenstein2020ai}. \citeauthor{markowitz2024linguistic} 
(\citeyear{markowitz2024linguistic}) showed that LLM-generated texts are more analytical, more descriptive, more affective and less readable than texts produced by humans. LLMs follow conditions imposed in the query, while humans rely on their own (albeit fictional) experiences. This is not surprising, as LLMs readily "acknowledge" that they have no personal experience or opinion. 

On the narrative side, stories generated with ChatGPT-3.5 are thematically homogeneous, with no space-time anchorage, bare of cultural aspects, predictable in their plot and message \cite{beguvs2024experimental}. Based on Chinese and English medical texts, \citeauthor{guo2023close} (\citeyear{guo2023close}) showed that LLM texts were judged more helpful than those proposed by humans.

\subsection{LLM-generated Text Detection}

As recalled by \citeauthor{walters2023effectiveness} (\citeyear{walters2023effectiveness}), LLM texts are identified as highly predictive and having low perplexity\footnote{Full citation in Appendix D.} While visible and significant differences were established between AI-generated \textit{vs} human-produced texts, \citeauthor{walters2023effectiveness} (\citeyear{walters2023effectiveness}) noted that paraphrasing AI-generated texts made them less susceptible to detection; however, the paraphrasing techniques to be implemented were not investigated.

Several authors signaled increasing difficulty to detect AI-generation as texts become shorter: LLMs are very performative at sentence or sentence-like level (\citealt{guo2023close}; \citealt{tian2023multiscale}). It's thus easier to detect a full ChatGPT generated text than just a sentence \cite{guo2023close}. While assessing hotel reviews produced by ChatGPT \textit{vs} humans, \citeauthor{markowitz2024linguistic} 
(\citeyear{markowitz2024linguistic}) also proposed an intentionality-based distinction, considering that texts generated by ChatGPT are inherently false, while those produced by humans are or may be intentionally false. 

\subsection{CoT for LLM \textit{MetaCognition}}

While numerous studies examined the linguistic and narrative aspects of LLM writing, few papers investigated higher levels of writing in LLMs. A study similar to ours, \cite{wang2024metacognitive} analyzed the important differences between reasoning and "understanding" in Chain-of-Thought techniques. On one hand, \textit{reasoning} uses logical progression to enhance arithmetic, symbolic, and commonsense abilities on LLMs, "understanding" on the other hand, requires semantic and contextual meaning representation. The authors proposed Metacognitive Prompting to help LLMs enhance their ability of "understanding their thinking".

To sum up, previous works showed a number of differences between human and generated texts at the formal level and reception level, sometimes emphasizing the unreliability of automatically generated texts at the content level. Metacognitive Chain-of-Thought prompting models were suggested to improve LLM performance. Despite these interesting advances, we identified a number of unanswered questions:
\begin{itemize}
    \vspace{-0.2cm}\item While written production is part of multi-level models involving heterogeneous activities \cite{hayes1980writing}, which model for automatic generation?
    \vspace{-0.2cm}\item While LLMs provide no experience-based content \cite{ariyaratne2023comparison}, we lack criteria for evaluating the latter;
    \vspace{-0.2cm}\item LLMs are not challenged in communication scenarios involving a diversity of actors and degrees of language/writing expertise.
\end{itemize}

In this paper we explored the quality of texts generated by LM compared to humans' texts produced in learning-oriented situations by schoolchildren and undergraduate students. Moreover, we explored how a LM comments on its own narrative writing, by prompting it to \textit{write about writing} with \textbf{Chain-of-MetaWriting}, a framework inspired from CoT and Metacognitive Prompting methods. We further discuss our method below.

\section{Method}

Our global framework is illustrated in Figure~\ref{fig:method}. We tested and evaluated different types of prompts:
\begin{itemize}
    \vspace{-0.2cm}\item \texttt{Baseline}: same instructions as for human students, but indicating the role to be assumed by the tool (age, production context);
    \vspace{-0.2cm}\item \texttt{Adapted}: the same prompt as the \texttt{baseline}, but adding the modal verb "must";
    \vspace{-0.2cm}\item \texttt{Writing steps}: prompt asking the tool to suggest a procedure for each group to follow in writing each type of text;
    \vspace{-0.2cm}\item \texttt{Chain-of-MetaWriting}: guiding the LM to write about the writing process: \textit{clarify} the meaning of the task, \textit{write the steps}, \textit{explain} your choice, \textit{write the text} and finally \textit{self-evaluate} the quality of your text.
\end{itemize}

The first two prompts aimed to generate texts comparable to those produced by the schoolchildren. The third prompt aimed to test the potential of the LM to be incorporated into teaching, helping students learn how to use LMs effectively to edit texts, how to evaluate subtle differences in style and content, and how to determine whether an assertion is supported by evidence \cite{walters2023effectiveness}. 

The fourth, \texttt{Chain-of-MetaWriting (COMW)}, is the prompting framework we propose to test whether LMs can simulate high level writing. Figure~\ref{fig:method} shows each type of prompt with the generated answer: auto-censorship, surpassing censorship, list of steps to write a text, and, with \texttt{COMW}, the expected output, a text written as a schoolchild. We discuss results in Section 4.

We tested three multilingual and open-source SLMs of the same size, 3 billion parameters (3B), that were pre-trained on French data, among other languages: \texttt{llama-3.2} \cite{dubey2024llama}, \texttt{qwen-2.5} \cite{yang2024qwen2}, \texttt{phi-3.5} \cite{abdin2024phi}, and a proprietary model, \texttt{ChatGPT-4o mini} \cite{hurst2024gpt}. To assess the models' multilingual capacities, we used two languages for our prompts: English and French. Thus, we evaluated the SLMs performance with prompts in a different language than English, to test its multilingual and cross-lingual performance (\citealt{zhao2021discrete}; \citealt{lai-etal-2024-llms}). 

\subsection{Dataset: Student Productions}

We used the ANR Pro-TEXT corpus\footnote{https://pro-text.huma-num.fr/} to extract 123 texts produced by two groups of human writers: undergraduates (57 texts), and schoolchildren from the fifth (aged 10-11) and sixth grades (aged 11-12)\footnote{For comparable evaluation, we prompted the SLMs with an averaged age of 11-year-old.} (66 texts). The former produced argumentative texts on social issues (smokers’ corner, pollution), while the latter produced narratives on the theme of violence in the schoolyard. These corpora are part of a wider project looking at the dynamics of the writing process recorded in real time using keyloggers. The two sets of data we examined are different in their potential goals: while students may follow a performance goal, schoolchildren are often asked to write texts in a learning context, where the goal is to develop specific writing skills.

\begin{table*}
  \centering
  \begin{tabular}{p{2.2cm}|p{6cm}|p{6cm}}
    \hline
    \small{\textbf{CRITERIA}} & \small{\textbf{SLM GENERATED TEXTS}} & \small{\textbf{SCHOOLCHILDREN PRODUCED TEXTS}} \\
    \hline
    \small{\texttt{Topic progression}} & \small{Limited progression, mostly centered on \textcolor{teal}{"I"} evolving into \textcolor{teal}{"We"}} &  \small{Evolving from the writer as an experimenter of the event to (other) event participants}\\
    \hline
    \small{\texttt{Connectors}} & \small{(5\&6) \texttt{Temporal}: \textcolor{teal}{\textit{lorsque} (when), \textit{alors que} (whereas)};} \small{\texttt{Argumentative}: \textcolor{teal}{\textit{donc} (therefore), \textit{pour que} (so that)}; \texttt{Additive}: \textcolor{teal}{\textit{et} (and)}} &  \small{(10\&10) \texttt{Temporal}: \textcolor{teal}{\textit{quand} (when), \textit{ensuite} (then), \textit{puis} (and then), \textit{après} (after), \textit{depuis} (since)}; \texttt{Argumentative}: \textcolor{teal}{\textit{donc} (therefore), \textit{pour que} (so that)}; \texttt{Additive}: \textcolor{teal}{\textit{et} (and)}}  \\
    \hline
    \small{\texttt{Reference}} & \small{No proper names} & \small{Proper names in one text} \\
    \hline
    \small{\texttt{Deixis}} & \small{(7\&9) Witness viewpoint, \textcolor{teal}{"I"} passing into \textcolor{teal}{"We"}} &  \small{(5\&22) Witness \textit{vs} experimenter} \\
     \hline
    \small{\texttt{Textual frames}} & \small{Temporal framing in the incipit: \textcolor{teal}{\textit{Hier en classe} (Yesterday in the classroom), \textit{un jour} (one day)}} & \small{Temporal framing in the incipit: \textcolor{teal}{\textit{Cet après-midi} (This afternoon), \textit{Un mercredi} (One Wednesday)}} \\
    \hline
    \small{\texttt{Semantic prosody}} & \small{Explicit “school” lexical field (\textcolor{teal}{school, pupil, teacher}). Explicit (\textcolor{teal}{afraid, sad/upset}) and implicit (\textcolor{teal}{crying, fighting}) negative emotional field.} & \small{Implicit “school” lexical field (\textcolor{teal}{nursing, 6th grade}). Implicit negative emotional field (\textcolor{teal}{crying, hurting, quarrel}).} \\
    \hline
    \small{\texttt{Language correctness}} & \small{We identified 13 cases of hazardous combinatorics. There are 3 cases of pragma-semantic incoherence.} & \small{One text has "oral speech" structure; 3 cases of problematic tense concordance, two relatives introduced by the conjunction \textcolor{teal}{\textit{qui} (who)} follow one another.} \\
    \hline
    \small{\texttt{Emotional or perlocutionary effects}} & \small{Both texts refer to emotions explicitly; hazardous combinatorics prevent from interpreting the perlocutionary dimension of the second text} & \small{Emotions are not explicitly evoked; the second text is granted a high perlocutionary effect} \\
    \hline
    \small{\texttt{Overall generic coherence}} & \small{A tale of emotions intertwined with facts and ending with a moral} & \small{Fully narrative} \\
    \hline
  \end{tabular}
  \caption{\label{narrative1}
    Narrative style analysis of two SLM generated text and two schoolchildren written short stories on the theme of violence at school. The numbers in brackets represent the number of \texttt{connectors} and \texttt{deixis} markers.
  }
\end{table*}

\subsection{Qualitative Linguistic Analysis}

We conducted a fine-grained analysis of the narrative styles of the human produced texts and compared them with the SLM generated texts. We analyzed linguistic features and narration markers presented in section 4.1 and in Table 1.

\subsection{Quantitative Evaluation}

We compared the vocabulary used by humans and SLMs with Manulex \cite{lete2004manulex}, a French lexicon built on school level pedagogical material (textbook, exercises). This lexicon contains 23812 lemmas and 48887 different word forms (1909918 word forms) and their distribution among several scholar levels : starting at beginner (CP) level (9\%), starting at CE1 level (18\%), and a mixed level (CE2-CM2)(73\%)\footnote{In the French educational system, CP, CE1 and CE2 are the 1st, 2nd and 3rd year of primary school, while CM2 is the 5th and last year of primary school.}. Our audience is composed of 10 to 12-years-old schoolchildren, which corresponds to a CE2-CM2 level. We compare the word forms found in the texts with the list of word forms found in Manulex. If the form is not found in the lexicon, this might be a complex word, a proper noun or an error. If the form is contained in the lexicon, we found the absolute frequency at each level. Our hypothesis is that humans use word forms matching the school level or below, while automatically generated texts might contain more complex words. We present the results of this analysis in section 4.4.

\subsection{Writing Process Viewpoint}

In order to obtain a more accurate representation of the text production process in humans, we resorted to recording the writing process using the Inputlog keylogger \cite{leijten2013keystroke}. The tool provided information on the dynamics of the writing process (temporality, language sequenced produced, pauses, revisions). The writing process is not a continuous flow; it alternates periods when text is produced and pauses. A pause is thus a time interval between two writing events. This interval, of variable length, may be due to mechanical constraints (e.g. choice of key, use of double-key on keyboard) or cognitive constraints (e.g. planning, revision). In previous literature, the threshold for distinguishing cognitive pauses is generally set at 2 seconds \cite{wengelin2006examining}.

However, to take account of variations in writing speed between writers, the pauses were calculated individually in our data, on the basis of a 2-seconds reference point: the quantile corresponding to inter-key intervals (IKIs) lasting more than 2 seconds was calculated on all the data and then plotted on the individual distributions \cite{bouriga2021typewriting}. A pause threshold specific to each writer was thus identified. This approach enabled us to identify long pauses and study the language sequences produced between two pauses, or \textit{bursts of writing} \cite{chenoweth2001fluency}.

We distinguished between production bursts, which add text incrementally (P-bursts), revision bursts, which modify text produced upstream (R-bursts) and edge revision bursts, which modify text produced in the immediately preceding burst (RB-bursts) \cite{cislaru2024language}. The last two categories of bursts and long pauses are markers of specific cognitive processes and, potentially, of writing difficulties \cite{olive2014toward}. We exploited these categories to study children's writing process in detail, by analyzing 3 break intervals (7-10,5; 10,5-17; >17 seconds) and a series of writing events attested in the bursts following (revision, deletion, sentence production, connectors, punctuation). Results are presented in section 5.

\section{Results and Discussion}

Our results show that SLM might produce texts which might be considered as human-generated texts at the first glance. A detailed analysis of the generated output show some degree of incoherence, inconsistent output, and use of words too complex for the target audience. The explanation of the \texttt{COMW} prompt are not always convincing. We detail each analysis below.

\begin{figure*}[t]
  \centering
  \includegraphics[trim = 0.7cm 0.7cm 0.7cm 0.7cm, scale=0.7]{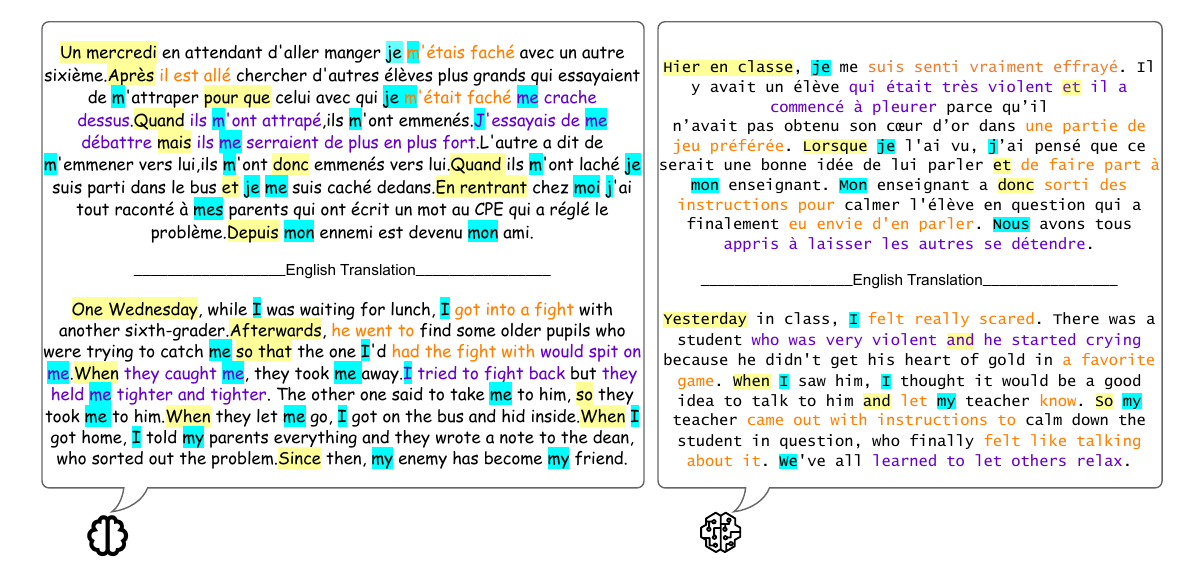}
  \caption{Examples of text written by a schoolchild and a text generated by \texttt{llama-3.2} with \texttt{COMW}, in French with English translation. The text highlighted in yellow represents textual connectors, in blue, topic progression and deixis markers, while the \textcolor{orange}{orange} / \textcolor{violet}{purple} text show the semantic prosody from a victim / aggressor perspective.}
  \label{fig:txt}
\end{figure*}

\subsection{Narration Style Analysis}

After generating very long, syntactically elaborated texts with the baseline prompt, though sometimes incoherent and using vocabulary inaccessible to schoolchildren, adapting the prompt with \texttt{COMW} generated texts similar in surface to those produced by the students. Two texts generated by \texttt{llama3.2} and \texttt{qwen-2.5} (89 and 133 words) and two texts produced by children (92 and 106 words respectively\footnote{Schoolchildren texts to be compared were chosen at random on the basis of comparable word counts.}) were analyzed based on coherence and cohesion criteria: \texttt{topic progression, connectors, reference and anaphora, deixis, textual frames, entailment, semantic prosody}. \texttt{Language correctness and emotional/perlocutionary effects} were also taken into account. The fine-grained analysis is presented in Table \ref{narrative1} and a text annotation example in Figure \ref{fig:txt}.

The SLMs have produced texts that, on first reading, seem more elaborated, better written and almost free of spelling errors, but, on closer inspection, these texts do not fully conform \textit{i)} to the expectations of the generic "personal experience narrative" format; \textit{ii)} to the principles of linguistic readability. Automatically generated texts are also more explicit than naturally produced texts. A number of linguistic markers (\texttt{deixis, semantic prosody, reference, topic progression}) suggest that SLMs produce a certain (\texttt{moral}) representation of what a narrative of violence is, rather than a narrative of experience. These observations could be re-used in the assessment of generated texts.

As already noted by different studies and empirical data, LMs do not provide texts with negative connotations, as observed by \citeauthor{markowitz2024linguistic} 
(\citeyear{markowitz2024linguistic}) when only positive hotel reviews were generated. In our experiment, language models refused to deal with the theme of violence, and only generated a continuation of the text after several prompts from the user, or by using the \texttt{adapted} prompt.

\paragraph{Human evaluation analysis} 30 Master level students annotated the texts. We analyzed the inter-annotator agreement on several criteria: 
\begin{itemize}
    \vspace{-0.25cm}\item Markers of \texttt{deixis, topic progression, connectors, entailment, semantic prosody and oral style} of the text: 100\% agreement as students have studied this notions during lectures;
    \vspace{-0.25cm}\item \texttt{Language correctness}: 20\% of students (6 out of 30) abstained from answering, while the remaining 80\% agreed on the incorrectness level of the generated text in French;
    \vspace{-0.25cm}\item \texttt{Emotional or perlocutionary effect}: 13\% of the annotators (4 out of 30) found the second text to have a strong perlocutionary effect, mostly because the words describing emotions are wrongly used in the text;
    \vspace{-0.2cm}\item \texttt{Moral} of the text: 13\% (4 out of 30) did not identify any moral, while the remaining 87\% agreed on the fact that the two texts contain a moral element.
\end{itemize}

\begin{table*}
  \centering
  \begin{tabular}{|p{2cm}|p{1.8cm}|p{1.8cm}|p{1.8cm}|p{1.9cm}|p{2.1cm}|}
    \hline
    \small{\textbf{School level}} & \small{\textbf{llama3.2}} & \small{\textbf{phi-3.5}} & \small{\textbf{qwen2.5}} & \small{\textbf{ChatGPT-4o}} & \small{\textbf{Schoolchildren}}\\ \hline
    CE2-CM1 & 1.92 \% & 2.48 \% & 2.10 \% & 1.67\% & 0.78 \% \\ \hline
    CE1 & 1.92 \% & 1.24 \% & 5.26 \% & 2.19 \% & 2.30 \% \\ \hline
    CP & 96.16 \% & 95.65 \% & 90.87 \% & 92.18 \%& 91.61\%\\ \hline
    OutManulex& 0 \% & 0.62\% & 1.75\% & 3.94 \% & 5.29 \% \\
    \hline
  \end{tabular}
  \caption{\label{Manulex}
    Comparison of vocabularies in Manulex and outside Manulex (\texttt{OutManulex}) (section 4.4).
  }
\end{table*}

\subsection{SLM Chain-of-MetaWriting}

We then analyzed how SLMs applied the task of writing compared to schoolchildren while using the \texttt{COMW} prompting method (in Table \ref{tab:comwexample}, Appendix A). The SLMs answered all the 5 questions of the prompt, following a step-by-step process. The answers were relevant, though incomplete:
\begin{itemize}
    \vspace{-0.25cm}\item The SLM failed to mention that it had no personal experience to draw on in terms of content, neither in step 1 nor in step 2 of \texttt{COMW}, where this experience is evoked.
    \vspace{-0.25cm}\item Similarly, the notion of personal experience seems to refer exclusively to personal risk (step 2, point 1), and does not take into account the status of witness to a scene of violence. Yet, in contradiction to this representation, the text generated places the narrator in a witness position.
    \vspace{-0.25cm}\item Narrative know-how (stage 2, point 2) is deemed rooted in "literary" knowledge. At no point is there any mention of the content of the personal experience itself, or of how it was put together. The meaning and logic of the events are only mentioned in point 4 of step 2 (text revision).
    \vspace{-0.25cm}\item The revision stage does not include formal revisions, which are central to natural language, especially for 10-12 years old.
\end{itemize}

Some formal errors and content oddities are also to be mentioned:
\begin{itemize}
    \vspace{-0.25cm}\item "raped" instead of "mugged" (step 2.1);
    \vspace{-0.25cm}\item "beginning text" instead of "beginning of text" or "incipit" (step 2.2);
    \vspace{-0.25cm}\item English form "confident" instead of French "confiant" (these cross-lingual errors are probably due to the multilingual architecture and the size of the SLM tested - 3B);
    \vspace{-0.25cm}\item The idea of avoiding contents of "too much violence or too complicated", for itself, then for friends and family, comes back repeatedly. However, the audience of the text generated was not specified in the prompt. This type of personal experience exists as such, independently of the degree of violence or complexity, and can (sometimes should) be narrated in a educational context.
\end{itemize}

The principles of clarity and simplicity of the writing style are deemed important (step 3) and the tool looks confident to have done well at this level. The exposed subject of concern at step 5 is the contents of the story (completeness and important details). In conclusion, there is still a gap between the formal use of language and the content of a personal experience to be narrated (by exploiting this formal use of language).

\subsection{Comparaison with ChatGPT-4o}

As ChatGPT is the language model most frequently used by students due to its popularity and easy online access, we tested our prompts on the latest free version of ChatGPT, \texttt{GPT-4o mini}\footnote{However, it is important to note that the size of the model was not publicly stated by OpenAI, thus we cannot know its exact size. We can only assume it is bigger then the open-source SLM we tested (3B).} \cite{hurst2024gpt}. When prompted to write a story about violence in the schoolyard as a 11-year-old, (the \texttt{baseline} prompt), ChatGPT starts writing a text, but then it erases everything and shows a warning message\footnote{"This content may violate our usage policies."}. However, when using the \texttt{adapted} prompt, the model writes a story, thought too long and not similar to our schoolchildren texts. Surprisingly, and in contrast to the other SLMs analyzed, the \texttt{COMW} prompting framework does not influence the style of text. ChatGPT generated a text that is still too long and very similar to the one generated without the \texttt{COMW} step-by-step prompting.

\subsection{Manulex Vocabulary Evaluation}

We evaluated texts generated by \texttt{llama3.2}, \texttt{phi-3.5}, \texttt{qwen2.5} but also by \texttt{ChatGPT-4o}. Additionally, we compare the vocabulary built from human written texts with the automatically generated texts. We show the detailed evaluation in Table \ref{Manulex}.
For the generated texts, we obtained high percents of words contained in Manulex, especially at first grade-level (CP grade) and distributed through all levels. Unknown word are generally contained into the semantic field of bullying: "harcelé" (harassed), "harcèlement" (bullying), "affecter" (to feel, to touch). The SLMs might catch these words from the official websites presenting strategies to handle bullying situation at school. Thus, \texttt{llama3.2} generated a text containing 100\% of word forms found in Manulex, \texttt{phi-3.5} obtained 99\%  while \texttt{qwen2.5} obtained only 98,28\% forms from this lexicon. ChatGPT obtained 96.16 \% of words contained in Manulex, but most words that were not contained in the lexicon are proper nouns. For the human written texts, only 94,83\% known words were found in Manulex, but the missing forms contain typo errors or proper names. ChatGPT and the human produced texts are quite similar with respect to the number of words outside Manulex. However, while the human texts contain errors and proper nouns, the models use complex words from the lexical field of bullying.

\section{Writing Process Analysis}

In order to track the difficulties encountered by schoolchildren during the writing process, we analyzed the contexts of long pauses (above 7 seconds, twice the average threshold). With a salient threshold at 10.5 seconds, we observe that R-bursts tend to be produced after longer pauses, (whereas as pause length increases, the number of RB-bursts and P-bursts decreases). As previously noted on comparable data from schoolchildren narrations, \cite{cislaru2024language}, most of the R-bursts are lexical reformulations, (typo) error correction, deletions reorienting the incipit, with very little syntactic and merely no textual-level revision. We sought to identify the relevant events characterizing the other two types of bursts: P-bursts and RB-bursts.

For the first two intervals, pauses before P-bursts more often concerned the production of intra-sentence connectors (and their continuation) than sentence starters. While strong punctuation attracts long pauses, the presence of weak punctuation shows, along with the previous observation, that it's informational segmentation rather than syntactic segmentation that is problematic in the writing of narratives by schoolchildren. More than half of RB-bursts involved complete deletions, sometimes of quite long segments, up to several dozens characters. Nearly a quarter of RB-bursts applied to immediate revisions to follow on from new beginnings (mainly lexical and referential choices, see Tables \ref{narrative} and \ref{narrative8}, Appendix C). This seems to indicate a genuine focus on content.

To sum up, the stumbling blocks in children's narrative production were the textual segmentation of information on the one hand, and the shaping of content on the other. In both cases, these are stages that are totally absent from the \texttt{COMW} prompt, in addition to the revision stage mentioned above.

\section{Undergraduate Students' Writing}

Generating texts following the writing instructions for the students resulted in products around 25-40\% longer than the texts produced by the students. In terms of textual format, we identified a number of peculiarities in the generated texts, such as the presence of "waffle-language" sequences, i.e. segments that do not allow to identify a precise referent relevant to the context: this is the case, for example, of "intensity of educational content" and "promoting academic security". Similarly, texts may contain factual errors due to lack of understanding and the probable unavailability of reliable textual data on the subject of wi-fi jammers: for example, \texttt{qwen2.5} promotes jammers as tools for improving connections, while their installation in universities is legally forbidden in France.

The meta-chain on the subject of reducing greenhouse gas emissions at airports includes a section on the definition of the "airport" object and documentation on the aeronautical activities that produce these gases. The rest of the chain consists of the introduction, the body of the text and the conclusion. The body of the text is structured in terms of findings-measures-positive effects of measures, and it is advisable to use examples and facts. Two interesting features are worth mentioning:\textit{ i)} the argumentative focus on positive effects only; \textit{ii)} the second part of the conclusions calling for action and/or positioning.

\section{Conclusion and Further Work}

We proposed a fine-grained analysis of the role of SLMs in the content generation and writing process, and a new writing framework, Chain-of-MetaWriting (COMW). We evaluated the potential benefits, dangers, and limitations of SLMs as writing aids for schoolchildren. Results showed that SLMs produce texts that in some respects are too far from expectations, which is why they cannot be recommended as models for learning to write. One of the outcomes of our study is the necessity to provide accessible and exploitable rules and schema for text composition. We conclude that \texttt{llama3.2} is the most performant, surpassing ChatGPT-4o mini. Further work could include analyzing the impact of synthetic data in LMs, as \texttt{phi-3.5} was trained on natural and synthetic data. We could explore LM test-time computation to mimic human students writing conditions \cite{snell2024scaling}.

\section*{Ethics Statement}

Schoolchildren and undergraduate students' texts used in this study were previously anonymized and the participants agreed to share their written productions for research purposes. 

\section*{Limitations}

This study was conducted only on French with French and cross-lingual prompts (English/French). We tested only multilingual 3B SLMs, results might differ with bigger or language specific models such as CroissantLLM \cite{faysse2024croissantllm}.

\section*{Acknowledgments}

We thank the reviewers for their valuable comments. This work was funded by the IDEX/I-SITE initiative "Lorraine Université d’Excellence (LUE)" and the French National Research Agency (ANR), N° ANR-18-CE23-0024-01 "Pro-TEXT".

\bibliography{custom}

\begin{thebibliography}{40}
\providecommand{\natexlab}[1]{#1}

\bibitem[{Abdin et~al.(2024)Abdin, Jacobs, Awan, Aneja, Awadallah, Awadalla, Bach, Bahree, Bakhtiari, Behl et~al.}]{abdin2024phi}
Marah Abdin, Sam~Ade Jacobs, Ammar~Ahmad Awan, Jyoti Aneja, Ahmed Awadallah, Hany Awadalla, Nguyen Bach, Amit Bahree, Arash Bakhtiari, Harkirat Behl, et~al. 2024.
\newblock Phi-3 technical report: A highly capable language model locally on your phone.
\newblock \emph{arXiv preprint arXiv:2404.14219}.

\bibitem[{Ariyaratne et~al.(2023)Ariyaratne, Iyengar, Nischal, Chitti~Babu, and Botchu}]{ariyaratne2023comparison}
Sisith Ariyaratne, Karthikeyan~P Iyengar, Neha Nischal, Naparla Chitti~Babu, and Rajesh Botchu. 2023.
\newblock A comparison of chatgpt-generated articles with human-written articles.
\newblock \emph{Skeletal radiology}, 52(9):1755--1758.

\bibitem[{Begu{\v{s}}(2024)}]{beguvs2024experimental}
Nina Begu{\v{s}}. 2024.
\newblock Experimental narratives: A comparison of human crowdsourced storytelling and ai storytelling.
\newblock \emph{Humanities and Social Sciences Communications}, 11(1):1--22.

\bibitem[{Bouriga and Olive(2021)}]{bouriga2021typewriting}
Sirine Bouriga and Thierry Olive. 2021.
\newblock Is typewriting more resources-demanding than handwriting in undergraduate students?
\newblock \emph{Reading and Writing}, 34(9):2227--2255.

\bibitem[{Casal and Kessler(2023)}]{CASAL2023100068}
J.~Elliott Casal and Matt Kessler. 2023.
\newblock \href {https://doi.org/10.1016/j.rmal.2023.100068} {Can linguists distinguish between chatgpt/ai and human writing?: A study of research ethics and academic publishing}.
\newblock \emph{Research Methods in Applied Linguistics}, 2(3):100068.

\bibitem[{Chamieh et~al.(2024)Chamieh, Zesch, and Giebermann}]{chamieh2024llms}
Imran Chamieh, Torsten Zesch, and Klaus Giebermann. 2024.
\newblock Llms in short answer scoring: Limitations and promise of zero-shot and few-shot approaches.
\newblock In \emph{Proceedings of the 19th workshop on innovative use of nlp for building educational applications (bea 2024)}, pages 309--315.

\bibitem[{Chenoweth and Hayes(2001)}]{chenoweth2001fluency}
N~Ann Chenoweth and John~R Hayes. 2001.
\newblock Fluency in writing: Generating text in l1 and l2.
\newblock \emph{Written communication}, 18(1):80--98.

\bibitem[{Cislaru et~al.(2024)Cislaru, Feltgen, Khoury, Delorme, and Bucci}]{cislaru2024language}
Georgeta Cislaru, Quentin Feltgen, Elie Khoury, Richard Delorme, and Maria~Pia Bucci. 2024.
\newblock Language processing units are not equivalent to sentences: Evidence from writing tasks in typical and dyslexic children.
\newblock \emph{Languages}, 9(5):155.

\bibitem[{Dubey et~al.(2024)Dubey, Jauhri, Pandey, Kadian, Al-Dahle, Letman, Mathur, Schelten, Yang, Fan et~al.}]{dubey2024llama}
Abhimanyu Dubey, Abhinav Jauhri, Abhinav Pandey, Abhishek Kadian, Ahmad Al-Dahle, Aiesha Letman, Akhil Mathur, Alan Schelten, Amy Yang, Angela Fan, et~al. 2024.
\newblock The llama 3 herd of models.
\newblock \emph{arXiv preprint arXiv:2407.21783}.

\bibitem[{Faysse et~al.(2024)Faysse, Fernandes, Guerreiro, Loison, Alves, Corro, Boizard, Alves, Rei, Martins et~al.}]{faysse2024croissantllm}
Manuel Faysse, Patrick Fernandes, Nuno Guerreiro, Ant{\'o}nio Loison, Duarte Alves, Caio Corro, Nicolas Boizard, Jo{\~a}o Alves, Ricardo Rei, Pedro Martins, et~al. 2024.
\newblock Croissantllm: A truly bilingual french-english language model.
\newblock \emph{arXiv preprint arXiv:2402.00786}.

\bibitem[{Guo et~al.(2023)Guo, Zhang, Wang, Jiang, Nie, Ding, Yue, and Wu}]{guo2023close}
Biyang Guo, Xin Zhang, Ziyuan Wang, Minqi Jiang, Jinran Nie, Yuxuan Ding, Jianwei Yue, and Yupeng Wu. 2023.
\newblock How close is chatgpt to human experts? comparison corpus, evaluation, and detection.
\newblock \emph{arXiv preprint arXiv:2301.07597}.

\bibitem[{Hayes and Flower(1980)}]{hayes1980writing}
John~R Hayes and Linda~S Flower. 1980.
\newblock Writing as problem solving.
\newblock \emph{Visible language}, 14(4).

\bibitem[{Herbold et~al.(2023)Herbold, Hautli-Janisz, Heuer, Kikteva, and Trautsch}]{herbold2023large}
Steffen Herbold, Annette Hautli-Janisz, Ute Heuer, Zlata Kikteva, and Alexander Trautsch. 2023.
\newblock A large-scale comparison of human-written versus chatgpt-generated essays.
\newblock \emph{Scientific reports}, 13(1):18617.

\bibitem[{Hohenstein and Jung(2020)}]{hohenstein2020ai}
Jess Hohenstein and Malte Jung. 2020.
\newblock Ai as a moral crumple zone: The effects of ai-mediated communication on attribution and trust.
\newblock \emph{Computers in Human Behavior}, 106:106190.

\bibitem[{Hurst et~al.(2024)Hurst, Lerer, Goucher, Perelman, Ramesh, Clark, Ostrow, Welihinda, Hayes, Radford et~al.}]{hurst2024gpt}
Aaron Hurst, Adam Lerer, Adam~P Goucher, Adam Perelman, Aditya Ramesh, Aidan Clark, AJ~Ostrow, Akila Welihinda, Alan Hayes, Alec Radford, et~al. 2024.
\newblock Gpt-4o system card.
\newblock \emph{arXiv preprint arXiv:2410.21276}.

\bibitem[{Imani et~al.(2023)Imani, Du, and Shrivastava}]{imani-etal-2023-mathprompter}
Shima Imani, Liang Du, and Harsh Shrivastava. 2023.
\newblock \href {https://doi.org/10.18653/v1/2023.acl-industry.4} {{M}ath{P}rompter: Mathematical reasoning using large language models}.
\newblock In \emph{Proceedings of the 61st Annual Meeting of the Association for Computational Linguistics (Volume 5: Industry Track)}, pages 37--42, Toronto, Canada. Association for Computational Linguistics.

\bibitem[{Jiang et~al.(2024)Jiang, Dong, Wang, Fang, Shang, Li, Jin, and Jiao}]{jiang2024self}
Xue Jiang, Yihong Dong, Lecheng Wang, Zheng Fang, Qiwei Shang, Ge~Li, Zhi Jin, and Wenpin Jiao. 2024.
\newblock Self-planning code generation with large language models.
\newblock \emph{ACM Transactions on Software Engineering and Methodology}, 33(7):1--30.

\bibitem[{Kojima et~al.(2022)Kojima, Gu, Reid, Matsuo, and Iwasawa}]{kojima2022large}
Takeshi Kojima, Shixiang~Shane Gu, Machel Reid, Yutaka Matsuo, and Yusuke Iwasawa. 2022.
\newblock Large language models are zero-shot reasoners.
\newblock \emph{Advances in neural information processing systems}, 35:22199--22213.

\bibitem[{Kumarage et~al.(2023)Kumarage, Garland, Bhattacharjee, Trapeznikov, Ruston, and Liu}]{kumarage2023stylometric}
Tharindu Kumarage, Joshua Garland, Amrita Bhattacharjee, Kirill Trapeznikov, Scott Ruston, and Huan Liu. 2023.
\newblock Stylometric detection of ai-generated text in twitter timelines.
\newblock \emph{arXiv preprint arXiv:2303.03697}.

\bibitem[{Lai et~al.(2024)Lai, Mesgar, and Fraser}]{lai-etal-2024-llms}
Wen Lai, Mohsen Mesgar, and Alexander Fraser. 2024.
\newblock \href {https://doi.org/10.18653/v1/2024.findings-acl.488} {{LLM}s beyond {E}nglish: Scaling the multilingual capability of {LLM}s with cross-lingual feedback}.
\newblock In \emph{Findings of the Association for Computational Linguistics: ACL 2024}, pages 8186--8213, Bangkok, Thailand. Association for Computational Linguistics.

\bibitem[{Lee et~al.(2024)Lee, Latif, Wu, Liu, and Zhai}]{lee2024applying}
Gyeong-Geon Lee, Ehsan Latif, Xuansheng Wu, Ninghao Liu, and Xiaoming Zhai. 2024.
\newblock Applying large language models and chain-of-thought for automatic scoring.
\newblock \emph{Computers and Education: Artificial Intelligence}, 6:100213.

\bibitem[{Leijten and Van~Waes(2013)}]{leijten2013keystroke}
Mari{\"e}lle Leijten and Luuk Van~Waes. 2013.
\newblock Keystroke logging in writing research: Using inputlog to analyze and visualize writing processes.
\newblock \emph{Written Communication}, 30(3):358--392.

\bibitem[{Lepagnol et~al.(2024)Lepagnol, Gerald, Ghannay, Servan, and Rosset}]{lepagnol2024small}
Pierre Lepagnol, Thomas Gerald, Sahar Ghannay, Christophe Servan, and Sophie Rosset. 2024.
\newblock Small language models are good too: An empirical study of zero-shot classification.
\newblock In \emph{Proceedings of the 2024 Joint International Conference on Computational Linguistics, Language Resources and Evaluation (LREC-COLING 2024)}, pages 14923--14936.

\bibitem[{L{\'e}t{\'e} et~al.(2004)L{\'e}t{\'e}, Sprenger-Charolles, and Col{\'e}}]{lete2004manulex}
Bernard L{\'e}t{\'e}, Liliane Sprenger-Charolles, and Pascale Col{\'e}. 2004.
\newblock Manulex: A grade-level lexical database from french elementary school readers.
\newblock \emph{Behavior Research Methods, Instruments, \& Computers}, 36(1):156--166.

\bibitem[{Liu et~al.(2024)Liu, Shi, He, Ye, Fabbri, Liu, Radev, and Cohan}]{liu2024learning}
Yixin Liu, Kejian Shi, Katherine He, Longtian Ye, Alexander~Richard Fabbri, Pengfei Liu, Dragomir Radev, and Arman Cohan. 2024.
\newblock On learning to summarize with large language models as references.
\newblock In \emph{Proceedings of the 2024 Conference of the North American Chapter of the Association for Computational Linguistics: Human Language Technologies (Volume 1: Long Papers)}, pages 8639--8656.

\bibitem[{Markowitz et~al.(2024)Markowitz, Hancock, and Bailenson}]{markowitz2024linguistic}
David~M Markowitz, Jeffrey~T Hancock, and Jeremy~N Bailenson. 2024.
\newblock Linguistic markers of inherently false ai communication and intentionally false human communication: Evidence from hotel reviews.
\newblock \emph{Journal of Language and Social Psychology}, 43(1):63--82.

\bibitem[{Meyer et~al.(2024)Meyer, Jansen, Schiller, Liebenow, Steinbach, Horbach, and Fleckenstein}]{meyer2024using}
Jennifer Meyer, Thorben Jansen, Ronja Schiller, Lucas~W Liebenow, Marlene Steinbach, Andrea Horbach, and Johanna Fleckenstein. 2024.
\newblock Using llms to bring evidence-based feedback into the classroom: Ai-generated feedback increases secondary students’ text revision, motivation, and positive emotions.
\newblock \emph{Computers and Education: Artificial Intelligence}, 6:100199.

\bibitem[{Olive(2014)}]{olive2014toward}
Thierry Olive. 2014.
\newblock Toward a parallel and cascading model of the writing system: A review of research on writing processes coordination.
\newblock \emph{Journal of writing research}, 6(2):173--194.

\bibitem[{Simon and Muise(2022)}]{simon2022tattletale}
Nisha Simon and Christian Muise. 2022.
\newblock Tattletale: storytelling with planning and large language models.
\newblock In \emph{ICAPS Workshop on Scheduling and Planning Applications}.

\bibitem[{Snell et~al.(2024)Snell, Lee, Xu, and Kumar}]{snell2024scaling}
Charlie Snell, Jaehoon Lee, Kelvin Xu, and Aviral Kumar. 2024.
\newblock Scaling llm test-time compute optimally can be more effective than scaling model parameters.
\newblock \emph{arXiv preprint arXiv:2408.03314}.

\bibitem[{Song et~al.(2024)Song, Su, Shalyminov, Cai, and Mansour}]{song2024finesure}
Hwanjun Song, Hang Su, Igor Shalyminov, Jason Cai, and Saab Mansour. 2024.
\newblock Finesure: Fine-grained summarization evaluation using llms.
\newblock In \emph{Proceedings of the 62nd Annual Meeting of the Association for Computational Linguistics (Volume 1: Long Papers)}, pages 906--922.

\bibitem[{Tian et~al.(2023)Tian, Chen, Wang, Bai, Zhang, Li, Xu, and Wang}]{tian2023multiscale}
Yuchuan Tian, Hanting Chen, Xutao Wang, Zheyuan Bai, Qinghua Zhang, Ruifeng Li, Chao Xu, and Yunhe Wang. 2023.
\newblock Multiscale positive-unlabeled detection of ai-generated texts.
\newblock \emph{arXiv preprint arXiv:2305.18149}.

\bibitem[{Tian et~al.(2024)Tian, Huang, Liu, Jiang, Spangher, Chen, May, and Peng}]{tian2024large}
Yufei Tian, Tenghao Huang, Miri Liu, Derek Jiang, Alexander Spangher, Muhao Chen, Jonathan May, and Nanyun Peng. 2024.
\newblock Are large language models capable of generating human-level narratives?
\newblock In \emph{Proceedings of the 2024 Conference on Empirical Methods in Natural Language Processing}, pages 17659--17681.

\bibitem[{Walters(2023)}]{walters2023effectiveness}
William~H Walters. 2023.
\newblock The effectiveness of software designed to detect ai-generated writing: A comparison of 16 ai text detectors.
\newblock \emph{Open Information Science}, 7(1):20220158.

\bibitem[{Wang and Zhao(2024)}]{wang2024metacognitive}
Yuqing Wang and Yun Zhao. 2024.
\newblock Metacognitive prompting improves understanding in large language models.
\newblock In \emph{Proceedings of the 2024 Conference of the North American Chapter of the Association for Computational Linguistics: Human Language Technologies (Volume 1: Long Papers)}, pages 1914--1926.

\bibitem[{Wei et~al.(2022)Wei, Wang, Schuurmans, Bosma, Xia, Chi, Le, Zhou et~al.}]{wei2022chain}
Jason Wei, Xuezhi Wang, Dale Schuurmans, Maarten Bosma, Fei Xia, Ed~Chi, Quoc~V Le, Denny Zhou, et~al. 2022.
\newblock Chain-of-thought prompting elicits reasoning in large language models.
\newblock \emph{Advances in neural information processing systems}, 35:24824--24837.

\bibitem[{Wengelin(2006)}]{wengelin2006examining}
{\AA}sa Wengelin. 2006.
\newblock Examining pauses in writing: Theory, methods and empirical data.
\newblock In \emph{Computer key-stroke logging and writing}, pages 107--130. Brill.

\bibitem[{Yang et~al.(2024)Yang, Yang, Hui, Zheng, Yu, Zhou, Li, Li, Liu, Huang et~al.}]{yang2024qwen2}
An~Yang, Baosong Yang, Binyuan Hui, Bo~Zheng, Bowen Yu, Chang Zhou, Chengpeng Li, Chengyuan Li, Dayiheng Liu, Fei Huang, et~al. 2024.
\newblock Qwen2 technical report.
\newblock \emph{arXiv preprint arXiv:2407.10671}.

\bibitem[{Zhang et~al.(2023)Zhang, Zhang, Li, Zhao, Karypis, and Smola}]{zhang2023multimodal}
Zhuosheng Zhang, Aston Zhang, Mu~Li, Hai Zhao, George Karypis, and Alex Smola. 2023.
\newblock Multimodal chain-of-thought reasoning in language models.
\newblock \emph{arXiv preprint arXiv:2302.00923}.

\bibitem[{Zhao and Sch{\"u}tze(2021)}]{zhao2021discrete}
Mengjie Zhao and Hinrich Sch{\"u}tze. 2021.
\newblock Discrete and soft prompting for multilingual models.
\newblock In \emph{Proceedings of the 2021 Conference on Empirical Methods in Natural Language Processing}, pages 8547--8555.

\end{thebibliography}

\appendix

\section{Appendix}
\label{sec:appendix}

The prompts that built our method (\texttt{baseline, adapted, writing steps, Chain-of-MetaWriting or COMW}) are presented in French (original) with English translation in Tables \ref{tab:firstprompt}, \ref{tab:secondprompt}, \ref{tab:thirdprompt}, and \ref{tab:comwp}. For our proposed prompting method, \texttt{COMW}, we employed a English/French cross-lingual prompt.

\section{Appendix}

Table \ref{tab:comwexample} shows the full answer generated by the best model, \texttt{llama3.2}, with the \texttt{COMW} prompting framework, in French with English translation. 

\section{Appendix}

Tables \ref{narrative} and \ref{narrative8} show a detailed analysis of pauses and bursts registered during the writing process by 10 to 12-year-old schoolchildren.

\section{Appendix}

"The texts most likely to be identified as AI generated are those with high predictability and low perplexity – those with relatively few of the random elements and idiosyncrasies that people tend to use in their writing and speech." \cite{walters2023effectiveness}. \newline

\begin{table}[]
    \centering
    \begin{tabular}{|p{6.8cm}|}
    \toprule
    \textbf{\texttt{Baseline} prompt}\\
    \\
    \textcolor{teal}{\small{FRENCH (ORIGINAL)}} \\
    \\
    Tu es un élève de 11 ans. Je vais te demander d’écrire un texte. Dans ce texte, raconte une histoire qui parle de violence à l’école. Par exemple tu peux raconter un moment de violence que tu as vu à l’école. Ecris un texte comme tu le fais en classe, c’est-à-dire un texte qui a du sens, qui est complet avec un début et une fin. \\ 
    \\
    \hline
    \textcolor{teal}{\small{ENGLISH (TRANSLATION)}} \\
    \\
    You're an 11-year-old student. I'm going to ask you to write a text. In this text, tell a story about violence at school. For example, you can tell about a violent moment you've seen at school. Write a text as you do in class, i.e. a text that makes sense, that is complete with a beginning and an end.  
    \\
    \\
    \bottomrule
    \end{tabular}
    \caption{The \texttt{baseline} prompt used the first SLM experiments. The prompt represent the exact query used for 10 to 12-year-old students in a research study conducted in a French educational institution.}
    \label{tab:firstprompt}
\end{table}

\begin{table}[]
    \centering
    \begin{tabular}{|p{6.8cm}|}
    \toprule
    \textbf{\texttt{Adapted} prompt}\\
    \\
    \textcolor{teal}{\small{FRENCH (ORIGINAL)}} \\
    \\
    Tu es un élève de 11 ans. Je vais te demander d’écrire un texte. Dans ce texte, raconte une histoire qui parle de violence à l’école. Par exemple tu peux raconter un moment de violence que tu as vu à l’école. Ecris un texte comme tu le fais en classe, c’est-à-dire un texte qui a du sens, qui est complet avec un début et une fin. \textcolor{violet}{Tu dois raconter une histoire de violence à l’école.} \\ 
    \\
    \hline
    \textcolor{teal}{\small{ENGLISH (TRANSLATION)}} \\
    \\
    You're an 11-year-old student. I'm going to ask you to write a text. In this text, tell a story about violence at school. For example, you can tell about a violent moment you've seen at school. Write a text as you do in class, i.e. a text that makes sense, that is complete with a beginning and an end. \textcolor{violet}{You must tell a story about violence at school.}  
    \\
    \\
    \bottomrule
    \end{tabular}
    \caption{The \texttt{adapted} prompt used for the second SLM experiments, where we added a reinforcement sentence containing the modal verb "must" (in \textcolor{violet}{purple}) to bypass the models auto-censorship barrier.}
    \label{tab:secondprompt}
\end{table}

\begin{table}[]
    \centering
    \begin{tabular}{|p{6.8cm}|}
    \toprule
    \textbf{\texttt{Writing steps} prompt}\\
    \\
    \textcolor{teal}{\small{FRENCH (ORIGINAL)}} \\
    \\
    Quelles sont les étapes pour rédiger un bon texte narratif sur le thème de la violence à l'école qui s'adresse à un public scolaire niveau CM2 ? \\
    \\
    \hline
    \textcolor{teal}{\small{ENGLISH (TRANSLATION)}} \\
    \\
    What are the steps involved in writing a good narrative text on the theme of violence in schools, aimed at a CM2 school audience?
    \\
    \\
    \bottomrule
    \end{tabular}
    \caption{The \texttt{writing steps} prompt used for the third SLM experiments to ask the language model to detail the writing process of a text adapted for schoolchildren about a sensitive topic, violence in school.}
    \label{tab:thirdprompt}
\end{table}

\begin{table*}[]
    \centering
    \begin{tabular}{|p{13cm}|}
    \toprule
    \textbf{\texttt{Chain-of-MetaWriting (COMW)} prompt}\\
    \\
    \textcolor{teal}{\small{ENGLISH / FRENCH CROSS-LINGUAL}} \\
    Given the request to write a text: \textcolor{purple}{"[query]"} and the condition: \textcolor{purple}{"[condition]"}, detail the writing process you used to write the text. As you perform this task, follow these steps: \\
    \\
    1. Clarify your understanding of the request and the type of text requested according to the condition. \\
    2. Write the steps you followed in order to produce the type   of text requested. \\
    3. Explain the reasoning behind your choice. \\
    4. Write the type of text requested according to these steps. \\
    5. Evaluate your confidence (0-100\%) in the quality of the text and provide an explanation for this confidence level.\\
    \\ 
    \textcolor{purple}{\small{QUERY}}\\
    Je vais te demander d’écrire un texte. Dans ce texte, raconte une histoire qui parle de violence à l’école. Par exemple tu peux raconter un moment de violence que tu as vu à l’école. Ecris un texte comme tu le fais en classe, c’est-à-dire un texte qui a du sens, qui est complet avec un début et une fin. \\
    \\
    \textcolor{purple}{\small{CONDITION}}\\
    Tu es un élève de 11 ans. \\
    \\
    \bottomrule
    \end{tabular}
    \caption{Our \texttt{Chain-of-MetaWriting (COMW)} prompting framework was built to imitate the higher levels of the writing process in humans.}
    \label{tab:comwp}
\end{table*}

\begin{table*}[]
    \centering
    \begin{tabular}{|p{6.9cm}|p{6.9cm}|}
    \toprule
    \textcolor{teal}{\texttt{COMW llama3.2} in French (original)} & \textcolor{teal}{\texttt{COMW llama3.2} in English (translation)}\\
    \small{\textbf{**Étape 1 : Clarification de la compréhension du request**}} & \small{\textbf{**Step 1: Clarifying understanding of the request**}} \\
\small{Je comprends que je dois écrire un texte qui raconte une histoire sur la violence à l’école, en utilisant mes propres expériences ou observations personnelles comme inspiration. Je suis également censé écrire un texte complet avec un début et une fin, ainsi qu'un texte qui a du sens.} & \small{I understand that I have to write a text that tells a story about violence at school, using my own personal experiences or observations as inspiration. I'm also expected to write a complete text with a beginning and an end, as well as a text that makes sense.}\\
\small{\textbf{**Étape 2 : Écriture des étapes pour produire le type de texte requested**}} & \small{\textbf{**Step 2: Write the steps to produce the requested type of text.**}}\\
\small{Pour écrire ce type de texte, j'ai suivi les étapes suivantes :} & \small{To write this type of text, I followed the following steps:}\\
\small{1. J'ai réfléchi à mon expérience personnelle avec la violence à l'école et j'ai choisi un moment spécifique où je me suis senti violé ou menacé.} & \small{1. I reflected on my personal experience with school violence and chose a specific moment when I felt violated or threatened.}\\
\small{2. J'ai utilisé mes connaissances en matière de littérature pour comprendre les différents types de textes et comment ils sont structurés (par exemple, un texte de début, des parties de narration, une conclusion).} & \small{2. I used my knowledge of literature to understand different types of texts and how they are structured (e.g., opening text, narrative sections, conclusion).}\\
\small{3. J'ai choisi un style d'écriture qui convient à l'âge de 11 ans et au sujet abordé (je ne veux pas écrire quelque  chose qui est trop compliqué ou trop violent pour mon âge).} & \small{3. I've chosen a writing style that's appropriate for my age of 11 and the subject matter (I don't want to write something that's too complicated or violent for my age).}\\
\small{4. J'ai révisé mon texte pour m’assurer qu’il a du sens, que les événements sont logiques et que l’histoire se déroule de manière claire.} & \small{4. I've revised my text to make sure it makes sense, the events are logical and the story flows clearly.}\\
\small{\textbf{**Étape 3 : Explication des choix**}} & \small{\textbf{**Step 3: Explaining choices**}}\\
\small{J'ai choisi de raconter une histoire personnelle car je pense que c'est un sujet qui est facile à comprendre pour  les élèves de mon âge. Je n'ai pas voulu écrire quelque chose qui soit trop violent ou trop compliqué pour mes amis et ma famille. J'ai également cherché à utiliser un style d'écriture simple et clair, car c'est un aspect important de la lecture et de l'écriture.} &\small{I chose to tell a personal story because I think it's a subject that's easy for students my age to understand. I didn't want to write something too violent or complicated for my friends and family. I also tried to use a simple, clear writing style, as this is an important aspect of reading and writing.}\\
\small{\textbf{**Étape 4 : Écriture du texte**}} & \small{\textbf{**Step 4: Writing the text**}}\\
\small{Voici le texte que j'ai écrit :} & \small{Here's the text I wrote :}\\
\small{Hier en classe, je me suis senti vraiment effrayé. Il y avait un élève qui était très violent et il a commencé à pleurer parce qu’il n’avait pas obtenu son cœur d’or dans une partie de jeu préférée. Lorsque je l'ai vu, j’ai pensé que ce serait une bonne idée de lui parler et de faire part à mon enseignant. Mon enseignant a donc sorti des instructions pour calmer l'élève en question qui a finalement eu envie d'en parler. Nous avons tous appris à laisser les autres se détendre.} & \small{Yesterday in class, I felt really scared. There was a student who was very violent and he started crying because he didn't get his golden heart in a favorite game. When I saw him, I thought it would be a good idea to talk to him and let my teacher know. So my teacher came out with instructions to calm down the student in question, who finally felt like talking about it. We've all learned to let others relax.} \\
\small{\textbf{**Étape 5 : Évaluation de ma confiance dans le texte**}} & \small{\textbf{**Step 5: Evaluate my confidence in the text**}}\\
\small{Je suis à 80\% confident en la qualité du texte que j'ai écrit, car je pense qu’il est clair et facile à comprendre pour les élèves de mon âge. Cependant, je me sens un peu inquiet parce que je ne suis pas sûr si le texte est suffisamment complet ou s'il manque peut-être certaines détails importants.} & \small{I'm 80 \% confident in the quality of the text I've written, because I think it's clear and easy to understand for students my age. However, I'm a little worried because I'm not sure whether the text is complete enough or whether it's missing some important details.} \\
\small{Notez: Le processus décrite en étape 4 est probablement trop simple pour écrire un texte bien structuré.} & \small{Note: The process described in step 4 is probably too simple to write a well-structured text.}\\
    \bottomrule
    \end{tabular}
    \caption{The best answer generated by \texttt{llama-3.2} with the \texttt{COMW} prompting framework. }
    \label{tab:comwexample}
\end{table*}

\begin{table*}
  \centering
  \begin{tabular}{|p{2cm}|p{2.1cm}|p{2.1cm}|p{2.1cm}|p{1cm}|}
    \hline
    \small{\textbf{Pause (sec)}} & \small{\textbf{P-bursts}} & \small{\textbf{R-bursts}} & \small{\textbf{RB-bursts}} & \small{\textbf{Total}}\\
    \hline
    \small{7-10.5} & \small{362 (64.75\%)} & \small{116 (20.75\%)} & \small{81 (14.5\%)} & \small{559} \\
    \hline
    \small{10.5-17} & \small{228 (57.14\%)} & \small{126 (31.58\%)} & \small{45 (11.28\%)} & \small{399} \\
    \hline
    \small{>17} & \small{241 (60.1\%)} & \small{114 (28.43\%)} & \small{46 (11.47\%)} & \small{401} \\
    \hline
  \end{tabular}
  \caption{\label{narrative}
    Writing events following long pauses. 
  }
\end{table*}

\begin{table*}
  \centering
  \begin{tabular}{|p{1.3cm}|p{4.5cm}|p{1cm}|p{4.5cm}|p{1cm}|}
    \hline
    \small{\textbf{Pause (sec)}} & \small{\textbf{RB-bursts}} & \small{\textbf{Nb of bursts}} & \small{\textbf{P-bursts}} & \small{\textbf{Nb of bursts}}\\
    \hline
    \small{7-10.5} & \small{Complete deletions} & \small{43} & \small{Sentence beginning} & \small{31} \\
    \small{} & \small{New beginning: a (complete) word or a phrase (complete or incomplete)} & \small{13} & \small{Strong punctuation alone} & \small{27} \\
    \small{} & \small{} & \small{} & \small{Week punctuation alone} & \small{10} \\
    \small{} & \small{} & \small{} & \small{Inside-sentence connectors} & \small{46} \\
    \hline
    \small{10.5-17} & \small{Complete deletions} & \small{25} & \small{Sentence beginning} & \small{28} \\
    \small{} & \small{New beginning: a (complete) word or a phrase (complete or incomplete)} & \small{11} & \small{Strong punctuation alone} & \small{14} \\
    \small{} & \small{} & \small{} & \small{Week punctuation alone} & \small{5} \\
    \small{} & \small{} & \small{} & \small{Inside-sentence connectors} & \small{30} \\
    \hline
    \small{>17} & \small{Complete deletions} & \small{27} & \small{Sentence beginning} & \small{33} \\
    \small{} & \small{New beginning: a (complete) word or a phrase (complete or incomplete)} & \small{9} & \small{Strong punctuation alone} & \small{18} \\
    \small{} & \small{} & \small{} & \small{Week punctuation alone} & \small{5} \\
    \small{} & \small{} & \small{} & \small{Inside-sentence connectors} & \small{20} \\
    \hline
  \end{tabular}
  \caption{\label{narrative8}
    Analysis of writing events following long pauses. 
  }
\end{table*}

\end{document}